\documentclass[runningheads]{llncs}
\usepackage{todonotes}
\usepackage{graphicx}
\usepackage{graphicx}
\usepackage{subfigure}
\usepackage{textcomp}
\usepackage{xcolor}
\usepackage{multirow}
\usepackage{enumitem}   
\usepackage{graphicx}
\usepackage{subfigure}
\usepackage{enumerate}
\usepackage{pdflscape}
\usepackage{booktabs}
\usepackage{url}
\usepackage{xcolor}
\usepackage{color}
\usepackage{array}
\usepackage{mathtools}
\usepackage{makeidx}
\usepackage{times}
\usepackage{amssymb}
\usepackage{amsmath}
\usepackage{inputenc}

\usepackage{epstopdf}
\usepackage{mathrsfs}
\usepackage{etoolbox}
\usepackage{wrapfig,lipsum,booktabs}
\begin{document}
\title{Cognitive Visual Commonsense Reasoning Using Dynamic Working Memory}

\author{Xuejiao Tang\inst{1}, Xin Huang\inst{2}, Wenbin Zhang\inst{2} \\Travers B. Child\inst{3}, Qiong Hu\inst{4}, Zhen Liu\inst{5}, and Ji Zhang\inst{6}}
\institute{$^1$Leibniz University of Hannover, Germany $^2$University of Maryland, Baltimore County, USA\\
$^3$China Europe International Business School, China $^4$Auburn University, USA\\
$^5$Guangdong Pharmaceutical University, China $^6$University of Southern Queensland, Australia\\
\email{\inst{1}xuejiao.tang@stud.uni-hannover.de, \inst{2}\{xinh1,wenbinzhang\}@umbc.edu\\ \inst{3}t.b.child@ceibs.edu, \inst{4}qzh0011@auburn.edu \\\inst{5}liu.zhen@gdpu.edu.cn, \inst{6}ji.zhang@usq.edu.au}}

\authorrunning{X. Tang, X. Huang, W. Zhang, T. B. Child, Q. Hu, Z. Liu, and J. Zhang}

\maketitle              
%
\begin{abstract}
\sloppy


Visual Commonsense Reasoning (VCR) predicts an answer with corresponding rationale, given a question-image input. VCR is a recently introduced visual scene understanding task with a wide range of applications, including visual question answering, automated vehicle systems, and clinical decision support. Previous approaches to solving the VCR task generally rely on pre-training or exploiting memory with long dependency relationship encoded models. However, these approaches suffer from a lack of generalizability and prior knowledge. In this paper we propose a dynamic working memory based cognitive VCR network, which stores accumulated commonsense between sentences to provide prior knowledge for inference. Extensive experiments show that the proposed model yields significant improvements over existing methods on the benchmark VCR dataset. Moreover, the proposed model provides intuitive interpretation into visual commonsense reasoning. A Python implementation of our mechanism is publicly available at \url{https://github.com/tanjatang/DMVCR} 
\end{abstract}

\section{Introduction}
Reflecting the success of Question Answering (QA)~\cite{hirschman2001natural} research in Natural Language Processing (NLP), many practical applications have appeared in daily life, such as Artificial Intelligence (AI) customer support, Siri, Alex, etc. However, the ideal AI application is a multimodal system integrating information from different sources~\cite{sharma2018conceptual}. For example, search engines may require more than just text, with image inputs also necessary to yield more comprehensive results. In this respect, researchers have begun to focus on multimodal learning which bridges vision and language processing. Multimodal learning has gained broad interest from the computer vision and natural language processing communities, resulting in the study of Visual Question Answering (VQA)~\cite{antol2015vqa}. VQA systems predict answers to language questions conditioned on an image or video. This is challenging for the visual system as often the answer does not directly refer to the image or video in question. Accordingly, high demand has arisen for AI models with cognition-level scene understanding of the real world. But presently, cognition-level scene understanding remains an open, challenging problem. To tackle this problem, Rowan Zeller et al.~\cite{zellers2019recognition} developed Visual Commonsense Reasoning (VCR). Given an image, a list of object regions, and a question, a VCR model answers the question and provides a rationale for its answer (both the answer and rationale are selected from a set of four candidates). As such, VCR needs not only to tackle the VQA task (i.e., to predict answers based on a given image and question), but also provides explanations for why the given answer is correct. VCR thus expands the VQA task, thereby improving cognition-level scene understanding. Effectively, the VCR task is more challenging as it requires high-level inference ability to predict rationales for a given scenario (i.e., it must infer deep-level meaning behind a scene).

The VCR task is challenging as it requires higher-order cognition and commonsense reasoning ability about the real world. For instance, looking at an image, the model needs to identify the objects of interest and potentially infer people's actions, mental states, professions, or intentions. This task can be relatively easy for human beings in most situations, but it remains challenging for up-to-date AI systems. Recently, many researchers have studied VCR tasks (see, e.g.,~\cite{zellers2019recognition,DBLP:journals/corr/Ben-younesCCT17,DBLP:journals/corr/abs-1910-14671,yu2020ernie,lu2019vilbert}). However, existing methods focus on designing reasoning modules without consideration of prior knowledge or pre-training the model on large scale datasets which lacks generalizability. To address the aforementioned challenges, we propose a Dynamic working Memory based cognitive Visual Commonsense Reasoning network (DMVCR), which aims to design a network mimicking human thinking by storing learned knowledge in a dictionary (with the dictionary regarded as prior knowledge for the network). In summary, our main contributions are as follows. First, we propose a new framework for VCR. Second, we design a dynamic working memory module with enhanced run-time inference for reasoning tasks. And third, we conduct a detailed experimental evaluation on the VCR dataset, demonstrating the effectiveness of our proposed DMVCR model.

The rest of this paper is organized as follows. In section~\ref{chap:relatedwork} we review related work on QA (and specifically on VCR). Section 3 briefly covers notation. In section~\ref{chap:framework} we detail how the VCR task is tackled with a dictionary, and how we train a dictionary to assist inference for reasoning. In section~\ref{chap:er} we apply our model to the VCR dataset. Finally, in section~\ref{chap:conclusion} we conclude our paper.

\section{Related Work}
\label{chap:relatedwork}
Question answering (QA) has become an increasingly important research theme in recent publications. Due to its broad range of applications in customer service and smart question answering, researchers have devised several QA tasks (e.g., Visual Question Answering (VQA)~\cite{antol2015vqa}, Question-Answer-Generation~\cite{lee2020generating}). Recently, a new QA task named
VCR~\cite{zellers2019recognition} provides answers with justifications for questions accompanied by an image. The key step in solving the VCR task is to achieve inference ability. There exists two major methods of enhancing inference ability. The first focuses on encoding the relationship between sentences using sequence-to-sequence based encoding methods. These methods infer rationales by encoding the long dependency relationship between sentences (see, e.g., R2C~\cite{zellers2019recognition} and TAB-VCR~\cite{DBLP:journals/corr/abs-1910-14671}). However, these models face difficulty reasoning with prior knowledge, and it is hard for them to infer reason based on commonsense about the world. The second method focuses on pre-training~\cite{yu2020ernie,chen2019uniter,lu2019vilbert}. Such studies typically leverage pre-training models on more than three other image-text datasets to learn various abilities like masked multimodal modeling, and multimodal alignment prediction~\cite{lu2019vilbert}. The approach then regards VCR as a downstream fine-tuning task. This method however lacks generalizability. 

Considering the disadvantages of either aforementioned approach, we design a network which provides prior knowledge to enhance inference ability for reasoning. The idea is borrowed from human beings' experience -- prior knowledge or commonsense provides rationale information when people infer a scene. To achieve this goal, we propose a working memory based dictionary module for run-time inference. Recent works such as~\cite{sabes1995advances,xiong2016dynamic,yang2019auto} have successfully applied the working memory into QA, VQA, and image caption. Working memory provides a dynamic knowledge base in these studies. However, existing work focuses on textual question answering tasks, paying less attention to inference ability~\cite{sabes1995advances,xiong2016dynamic}. Concretely, the DMN network proposed in~\cite{kumar2016ask} uses working memory to predict answers based on given textual information. This constitutes a step forward in demonstrating the power of dynamic memory in QA tasks. However, that approach can only tackle textual QA tasks. Another work in~\cite{xiong2016dynamic} improves upon DMN by adding an input fusion layer (containing textual and visual information) to be used in VQA tasks. However, both methods failed to prove the inference ability of dynamic working memory. Our paper proposes a dictionary unit based on dynamic working memory to store commonsense as prior knowledge for inference.  
\section{Notations and Problem Formulation}
\label{chap:npf}
The VCR dataset consists of millions of labeled subsets. Each subset is composed of an image with one to three associated questions. Each question is then associated with four candidate answers and four candidate rationales. The overarching task is formulated as three subtasks: (1) predicting the correct answer for a given question and image ($Q \rightarrow A$); (2) predicting the correct rationale for a given question, image, and correct answer ($QA \rightarrow R$); and (3) predicting the correct answer and rationale for a given image and question ($Q \rightarrow AR$). Additionally, we defined two language inputs - query $q\{q_1,q_2,\cdots,q_n\}$ and response $r\{r_1,r_2,\cdots,r_n\}$, as reflected in Figure~\ref{fig:framework}. In the $Q \rightarrow A$ subtask, query $q$ is the question and response $r$ is the answers. In the $QA \rightarrow R$ subtask, query $q$ becomes the question together with correct answer, while rationales constitute the response $r$. 

\section{Proposed Framework}
\label{chap:framework}
\begin{figure}[!htbp]
	\vspace{-2.8cm}
	\centering
	\includegraphics[width=\textwidth]{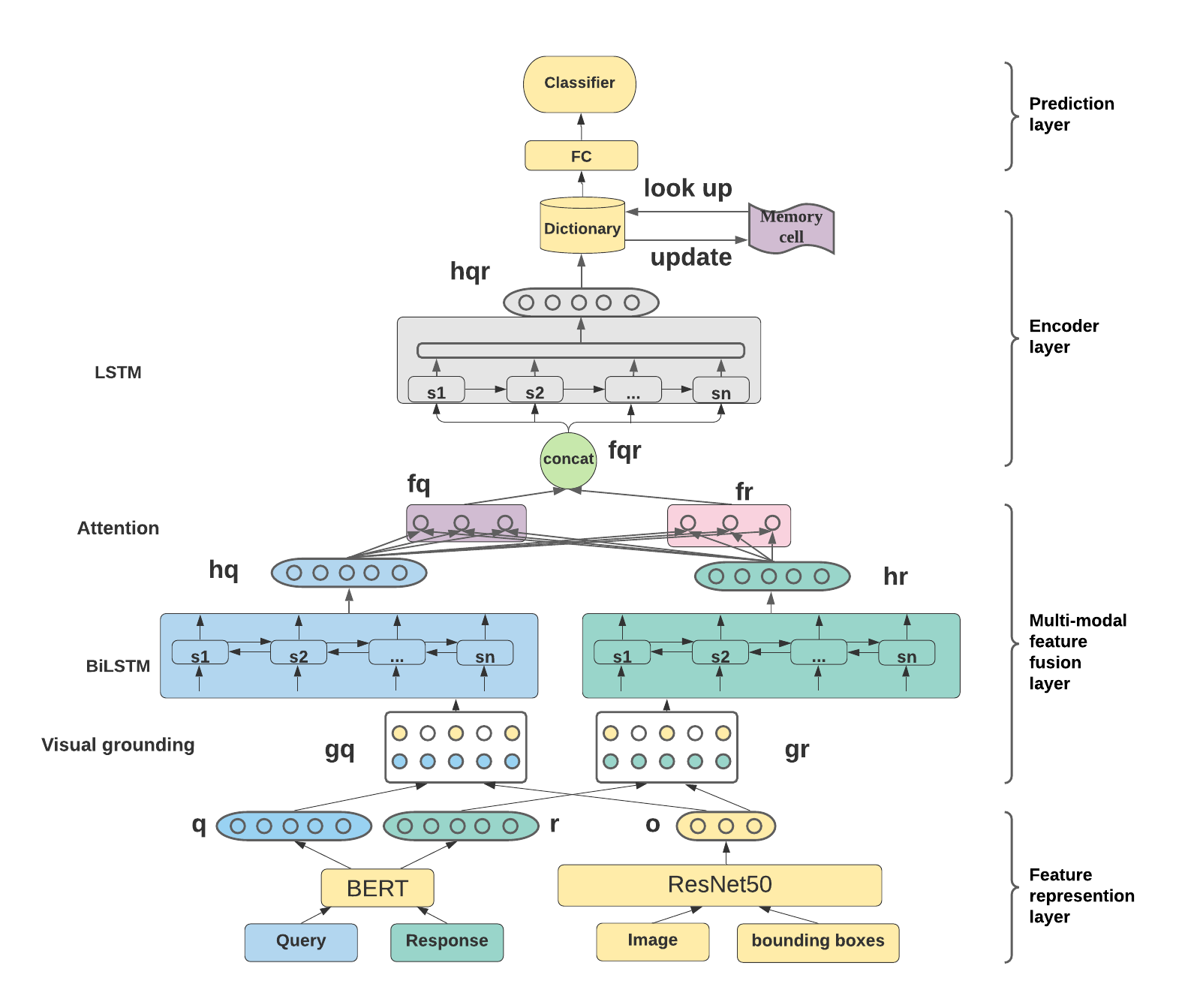}
	\caption{High-level overview of the proposed DMVCR consisting of four modules: feature representation layer to extract visual and textual features; multimodal feature fusion to contextualize multimodal representations; encoder layer to encode rich visual commonsense; and prediction layer to select the most related response.}
	\label{fig:framework}
\end{figure}

As shown in Figure~\ref{fig:framework}, our framework consists of four layers: a feature representation layer, a multimodal fusion layer, an encoder layer, and a prediction layer. The first layer captures language and image features, and converts them into dense representations. The represented features are then fed into the multimodal fusion layer to generate meaningful contexts of language-image fused information. Next, the fused features are fed into an encoder layer, which consists of a long dependency encoder~\cite{huang2015bidirectional} RNN module along with a dictionary unit. Finally, a prediction layer is designed to predict the correct answer or rationale.

\subsection{Feature Representation Layer}
\label{frl}
The feature representation layer converts features from images and language into dense representations. For the language, we learn embeddings for the query $q\{q_1,q_2,\cdots,q_n\}$ and response $r\{r_1,r_2,\cdots,r_n\}$ features. Additionally,
the object features $o\{o_1,o_2,\cdots,\\o_n\}$ are extracted from a deep network based on residual learning~\cite{he2016deep}.

\subsubsection*{Language embedding.} 
\label{le}
The language embeddings are obtained by transforming raw input sentences into low-dimensional embeddings. The query represented by $q\{q_1,q_2,\\\cdots,q_n\}$ refers to a question in the question answering task ($Q \rightarrow A$), and a question paired with correct answer in the reasononing task ($QA \rightarrow R$). Responses $r\{r_1,r_2,\cdots,\\r_n\}$ refer to answer candidates in the question answering task ($Q \rightarrow A$), and rationale candidates in the reasoning task ($QA \rightarrow R$). 
The embeddings are extracted using an attention mechanism with parallel structure~\cite{devlin2018bert}. Note that the sentences contain tags related to objects in the image. For example, see Figure~\ref{fig:Qualitative result example 1} and the question ``Are [0,1] happy to be here?'' The [0,1] are tags set to identify objects in the image (i.e., the object features of person 1 and person 2).


\subsubsection*{Object embedding.} The images are filtered from movie clips. To ensure images with rich information, a filter is set to select images with more than two objects each~\cite{zellers2019recognition}. The  object features are then extracted with a residual connected deep network~\cite{he2016deep}. The output of the deep network is object features with low-dimensional embeddings $o \{o_1,o_2,\cdots,o_n\}$.

\subsection{Multimodal Feature Fusion Layer} The multimodal feature fusion layer consists of three modules: a visual grounding module, an RNN module, and an attention module. 

\subsubsection*{Visual grounding.} Visual grounding aims at finding out target objects for query and response in images. As mentioned in section~\ref{le}, tags are set in query and responses to reference corresponding objects. The object features will be extracted and concatenated to language features at the visual grounding unit to obtain the representations with both image and language information. As shown in Figure~\ref{fig:framework}, the inputs of visual grounding consist of language ($q\{ q_1,q_2,\cdots,q_n\}$ and $r\{r_1,r_2,\cdots,r_n \}$) along with related objects features ($o \{o_1,o_2,\cdots,o_m\}$). The output contains aligned language and objects features ($g_q$ and $g_r$). The white unit at visual grounding is grounded representations, which contains image and text information. It can be formulated as follows (where $concat$ represents the concatenate operation):
\begin{align}
    g_r = concat(o, r)\\
    g_q = concat(o, q)
\end{align}

\subsubsection*{RNN module.} The grounded language and objects features $g_q$ and $g_r$ at the visual grounding stage contain multimodal information from images and text. However, they cannot understand the semantic dependency relationship around each word. To obtain language-objects mixed vectors with rich dependency relationship information, we feed the aligned language features $g_q$ and $g_r$ into BiLSTM~\cite{huang2015bidirectional}, which exploits the contexts from both past and future. In details, it increases the amount of information by means of two LSTMs, one taking the input in a forward direction with hidden layer $\overrightarrow{h_{lt}}$, and the other in a backwards direction with hidden layer $\overleftarrow{h_{lt}}$. The query-objects representations and response-objects $\underset{l \in \{q,r\}}{h_l}$ output at each time step $t$ is formulated as:

\begin{align}
\label{bih}
    h_{lt} =  \overrightarrow{h_{lt}} \oplus \overleftarrow{h_{lt}}\\
	\overrightarrow{h_{lt}} = o_t \odot \tanh(c_t)
\end{align}


\noindent where $c_t$ is the current cell state and formulated as:
\begin{align}
\label{lstm}
    c_t = f_t \odot c_{t-1} + i_t \odot tanh(W_c \cdot [c_{t-1}, h_{l(t-1)}, x_t] + b_c)\\
	i_t = \sigma(W_i \cdot [c_{t-1}, h_{l(t-1)}, x_t] + b_i)\\
	o_t = \sigma(W_o \cdot [c_{t-1}, h_{l(t-1)}, x_t] + b_o)\\
	f_t = \sigma(W_f \cdot [c_{t-1}, h_{l(t-1)}, x_t] + b_f)
\end{align}

\noindent where $i, o, f$ represent input gate, output gate, and forget gate, respectively, and $x_t$ is the $t^{th}$ input of a senquence. In addition, $W_i, W_o, W_f, W_c, b_c, b_i, b_o, b_f$ are trainable parameters with $\sigma$ representing the sigmoid activation function~\cite{yin2003flexible}.


\subsubsection*{Attention module.}
Despite the good learning of BiLSTM in modeling sequential transition patterns, it is unable to fully capture all information from images and languages. Therefore, an attention module is introduced to enhance the RNN module, picking up object features which are ignored in the visual grounding and RNN modules. The attention mechanism on object features $o \{o_1,o_2,\cdots,o_n\}$ and response-objects representations $h_r$ is formulated as:

\begin{align}
\alpha_{i,j} = softmax({o_iW_rh_{rj}})\\
\hat{fr_i} = \sum_j \alpha_{i,j}h_{rj}
\end{align}
\noindent where $i$ and $j$ represent the position in a sentence, $W_r$ is trainable weight matrix. In addition, this attention step also contextualizes the text through object information. 

Furthermore, another attention module is implemented between query-objects representations $h_q$ and response-objects representations $h_r$, so that the output fused query-objects representation contains weighted information of response-objects representations.  It can be formulated as:

 \begin{align}
	 \alpha_{i,j} = softmax({h_{ri}W_qh_{qj}})\\
	 \hat{fq_i} = \sum_j \alpha_{i,j}h_{qj}
 \end{align}
\noindent where $W_q$ is the trainable weight matrix, $i$ and $j$ denote positions in a sentence.

\subsection{Encoder Layer}
The encoder layer aims to capture the commonsense between sentences and use it to enhance inference.
It is composed of an RNN module and a dictionary module.

\subsubsection*{RNN module.}
An RNN unit encodes the fused queries and responses by long dependency memory cells~\cite{Hochreiter1997LongSM}, so that relationships between sentences can be captured. The input is fused query ($\hat{f_q}$) and response ($\hat{f_r}$) features. To encode the relationship between sentences, we concatenate $\hat{f_q}$ and $\hat{f_r}$ at sentence length dimensions as the input of LSTM. Its last output hidden layer contains rich information about commonsense between sentences. At time step $t$, the outputting representations can be formulated as: 
\begin{align}
	h_t = o_t \odot tanh(c_t)
\end{align}

\noindent where the $c_t$ is formulated the same as in Equation~(\ref{lstm}). The difference is that $x_{t} = concat(\hat{f_q}, \hat{f_q})$, where concat is the concatenate operation. In addition, the outputting representations $h_t$ is the last hidden layer of LSTM, while the outputting in Equation~(\ref{bih}) is every time step of BiLSTM.


\subsubsection*{Dictionary module.}
Despite effective learning of the RNN unit in modeling the relationship between sentences, it is still limited for run-time inference. We therefore propose a dictionary unit to learn dictionary $D$, and then use it to look up commonsense for inference. The dictionary is a dynamic knowledge base and is being updated during training. We denote the dictionary as a $d \times k$ matrix $D \{d_1, d_2,..., d_k\}$, where $k$ is the size of dictionary. The given encoded representation $h$ from RNN module will be encoded using the formulations:

\begin{align}
\hat{h} = \sum_{k=1}^K \alpha_kd_k, \alpha = softmax(D^Th)
\end{align} 
\noindent where $\alpha$ can be viewed as the ``key'' idea in memory network~\cite{yang2019auto}.

\subsection{Prediction Layer}
The prediction layer generates a probability distribution of responses from the high-dimension context generated in the encoder layer. It consists of a multi-layer perceptron. VCR is a multi-classification task in which one of the four responses is correct. Therefore, multiclass
cross-entropy~\cite{rubinstein1999cross} is applied to complete the prediction.

\section{Experimental Results}
\label{chap:er}
In this section, we conduct extensive experiments to demonstrate the effectiveness of our proposed DMVCR network for solving VCR tasks. We first introduce the datasets, baseline models, and evaluation metrics of our experiments. Then we compare our model with baseline models, and present an analysis of the impact of the different strategies. Finally, we present an intuitive interpretation of the prediction.

\subsection{Experimental Settings}
\subsubsection*{Dataset.} 
The VCR dataset~\cite{zellers2019recognition} is composed of 290k multiple-choice questions in total (including 290k correct answers,
290k correct rationales, and 110k images). The correct answers and rationales labeled in the dataset are met with 90\% of human agreements. An adversarial matching approach is adopted to obtain counterfactual choices with minimal bias. Each answer contains 7.5 words on average, and each rationale contains 16 words on average. Each set consists of an image, a question, four available answer choices, and four candidate rationales. The correct answer and rationale are given in the dataset.

\begin{figure}[!htbp]
    \centering
    \includegraphics[width=7cm]{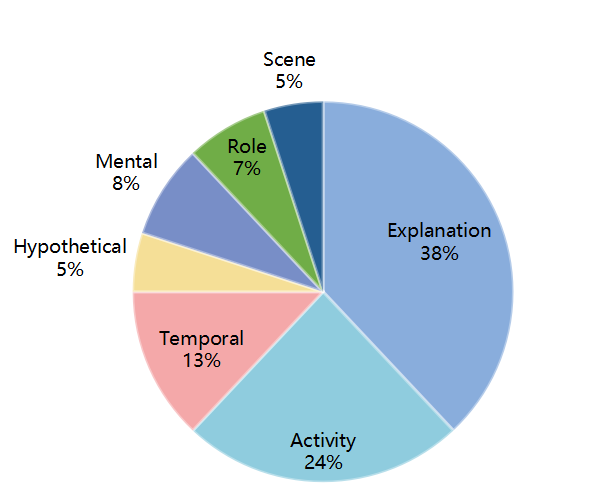}
    \caption{Overview of the types of inference required by questions in VCR.}
    \label{fig:Datadis}
\end{figure}

The distribution of inference types is shown in Figure~\ref{fig:Datadis}. Thirty-eight percent of the inference types are regarding explanation, and 24\% of them are about the activity. The rest are related to temporal,  mental, role, scene, and hypothetical inference problems.
 
\subsubsection*{Hyperparameters.}
The image features are projected to 512 dimension. The word embedding dimension is 768. The dictionary is a [512,800] matrix, where 512 is the embedding dimension, and 800 is the dictionary size. We separately set the learning rate for the memory cell (the dictionary cell) to 0.02, and others to 0.0002. In addition, for the $Q \rightarrow A$ subtask, we set the hidden size of LSTM encoder to 512. For the $QA \rightarrow R$ subtask, we set the hidden size of LSTM encoder to 64. The model was trained with the Adam algorithm~\cite{kingma2014adam} using PyTorch on NVIDIA GPU GTX 1080.


\subsubsection*{Metric.}
The VCR task can be regarded as a multi-classification
problem. We use mAp~\cite{henderson2016end} to evaluate the performance, which is a common metric for evaluating prediction accuracy in multi-classification areas.

\subsubsection*{Approach for comparison.}
We compare the proposed DMVCR with recent deep learning-based models for VCR. Specifically, the following baseline approaches are evaluated:
\begin{itemize}
    \item \textbf{RevisitedVQA~\cite{jabri2016revisiting}:}
    Different from the recently proposed systems, which have a reasoning module that includes an attention mechanism or memory mechanism, RevisitedVQA focuses on developing a ``simple'' alternative model, which reasons the response using logistic regressions and multi-layer perceptrons (MLP).
     \item \textbf{BottomUpTopDown~\cite{anderson2018bottom}:} Proposed a bottom-up and top-down attention method to determine the feature weightings for prediction. It computes a weighted sum over image locations to fuse image and language information so that the model can predict the answer based on a given scene and question.
     \item \textbf{MLB~\cite{kim2016hadamard}:} 
     Proposed a low-rank bilinear pooling for the task. The bilinear pooling is realized by using the Hadamard product for attention mechanism and has two linear mappings without
biases for embedding input vectors.
     \item \textbf{MUTAN~\cite{ben2017mutan}:}
     Proposed a multimodal fusion module tucker decomposition (a 3-way tensor), to fuse image and language information. In addition, multimodal low-rank bilinear (MLB) is used to reason the response for the input.
     \item \textbf{R2C~\cite{zellers2019recognition}:}
     Proposed a fusion module, a contextualization module, and a reasoning module for VCR. It is based on the sequence relationship model LSTM and attention mechanism. 
\end{itemize}

\subsection{Analysis of Experimental Results}

\subsubsection*{Task description.}
We implement the experiments separately in three steps. We firstly conducted $Q \rightarrow A$ evaluation, and then $QA \rightarrow R$. Finally, we join the $Q \rightarrow A$ result and $QA \rightarrow R$ results to obtain the final $Q \rightarrow AR$
prediction result. The difference between the implementation of $Q \rightarrow A$ and $QA \rightarrow R$ tasks is the input query and
response. For the $Q \rightarrow A$ task, the query is the paired question, image, four candidate answers; while the response is the correct answer. For the $QA \rightarrow R$ task, the query is the paired question, image, correct answer, and four candidate rationales; while the response is the correct rationale.
\begin{table}[!htpb]
\centering
\renewcommand\arraystretch{1.2}
\begin{tabular}{|p{4cm}<{\centering}|p{2cm}<{\centering}|p{2cm}<{\centering}|p{2cm}<{\centering}|}
\hline
\textbf{Models} &\textbf{$Q \rightarrow A$} &\textbf{$QA \rightarrow R$} &\textbf{$Q \rightarrow AR$} \\
\hline
\textbf{RevisitedVQA~\cite{jabri2016revisiting}} & 39.4 & 34.0 & 13.5 \\
\hline
\textbf{BottomUpTopDown~\cite{anderson2018bottom}} & 42.8 & 25.1 & 10.7 \\
\hline
\textbf{MLB~\cite{kim2016hadamard}} & 45.5 & 36.1 & 17\\
\hline
\textbf{MUTAN~\cite{ben2017mutan}} & 44.4 & 32.0 & 14.6\\
\hline
\textbf{R2C (Baseline)~\cite{zellers2019recognition}} & \textbf{61.9} & \textbf{62.8} & \textbf{39.1}\\
\hline
\textbf{DMVCR} & \textbf{62.4} (+0.8\%) & \textbf{67.5} (+7.5\%) & \textbf{42.3} (+8.2\%)\\
\hline
\end{tabular}
\caption{Comparison of results between our methods and other popular methods using the VCR Dataset. The best performance of the compared methods is highlighted. Percentage in parenthesis is our relative improvement over the performance of the best baseline method.}
\label{tab:experment results}
\end{table}
\subsubsection*{Analysis.}
We evaluated our method on the VCR dataset and compared the performance with other
popular models. As the results in Table~\ref{tab:experment results} show, our approach outperforms in all of the subtasks: $Q \rightarrow A$, $QA \rightarrow R$, and $Q \rightarrow AR$. Specifically, our method outperforms MUTAN and MLB by a large margin. Furthermore, it also performs better than R2C. 

\begin{figure}[]
 \label{fig:Qualitative results}
 \addtocounter{subfigure}{0} 
 \vspace{-0.5cm}
 \centering
 \subfigure[Qualitative example 1. The model predicts the correct answer and rationale.]{
 \label{fig:Qualitative result example 1}
 \includegraphics[width = 9.35cm,height=6.3cm]{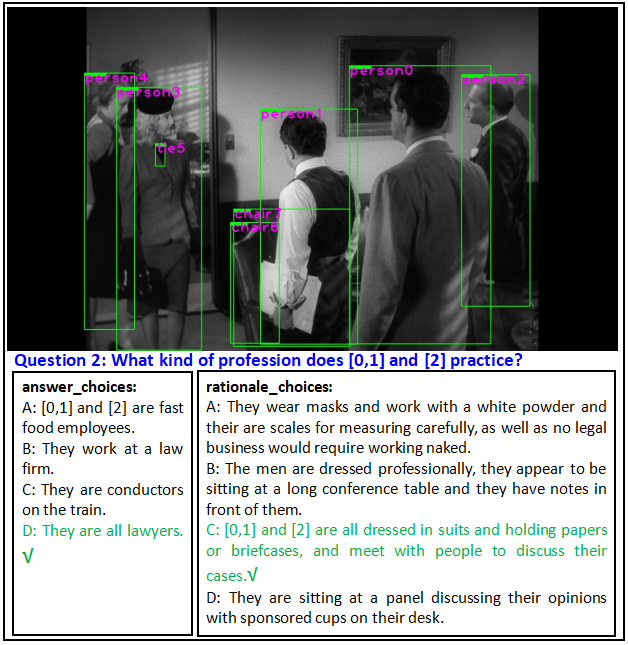}
 }
 \subfigure[Qualitative example 2. The model predicts the correct answer and rationale.]{
 \label{fig:Qualitative result example 2}
     \includegraphics[width = 9.35cm,height=6.7cm]{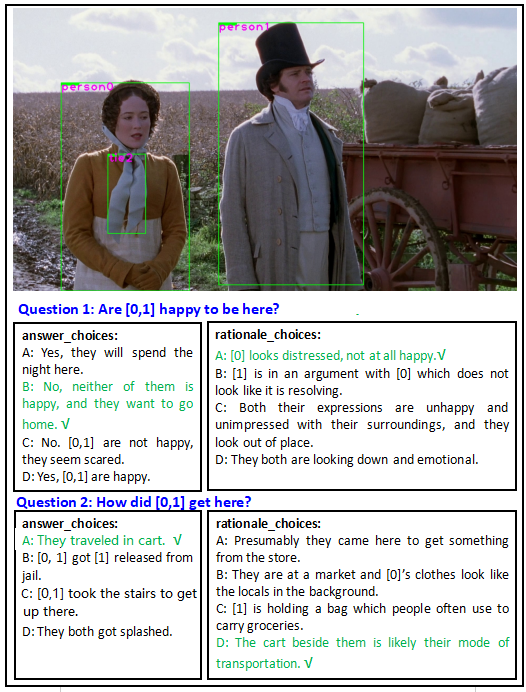}}
 \subfigure[Qualitative example 3. The model predicts the correct answer but incorrect rationale in Question 1. The model predicts an incorrect answer but correct rationale in Question 2.]{
 \label{fig:Qualitative result example 3}
     \includegraphics[width = 9.35cm,height=6.7cm]{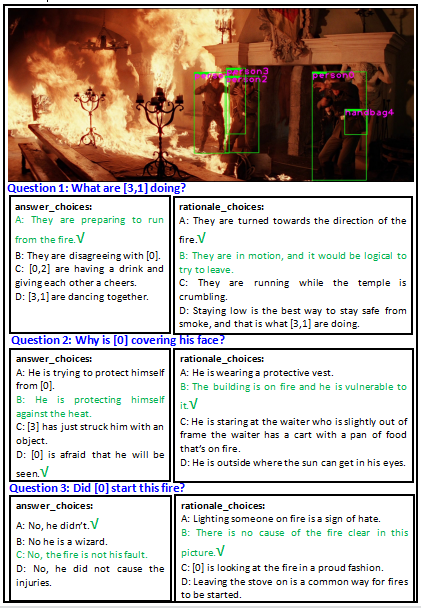}}
\vspace{-0.4cm}
 \caption{Qualitative examples. Prediction from DMVCR is marked with a \textcolor{green}{\checkmark} while correct results are highlighted in green.}
 \end{figure}

\subsection{Qualitative Results}

We evaluate qualitative results on the DMVCR model. The qualitative examples are provided in
Figure 3. The candidate in green represents the correct choice; the candidate with a checkmark \textcolor{green}{\checkmark} represents the prediction result by our proposed DMVCR model.
As the qualitative results show, the DMVCR model improves its power in inference. 
 
For instance, see Figure~\ref{fig:Qualitative result example 1}. The question listed is: ``What kind of profession does [0,1] and [2] practice?''. The predicted answer is D - ``They are all lawyers.'' Furthermore, the model offers rationale C - ``[0,1] and [2] are all dressed in suits and holding papers or briefcases, and meet with people to discuss their cases.'' DMVCR correctly infers the rationale based on dress and activity, even though this task is difficult for humans. 

DMVCR can also identify human beings' expressions and infer emotion. See for example the result in Figure~\ref{fig:Qualitative result example 2}. Question 1 is: ``Are [0,1] happy to be there?''. Our model selects the correct answer B along with reason A: ``No, neither of them is happy, and they want to go home''; because ``[0] looks distressed, not at all happy.'' 

Finally, there are also results which predict the correct answer but infer the wrong reason. For instance, see question 1 in Figure~\ref{fig:Qualitative result example 3}: ``What are [3,1] doing?'' DMVCR predicts the correct answer A - ``They are preparing to run from the fire.'' But it infers the wrong reason A - ``They are turned towards the direction of the fire.'' The correct answer is of course B - ``They are in motion, and it would be logistical to try to leave.'' It is also possible for the model to predict a wrong answer but correct rationale. This appears in question 2 of Figure~\ref{fig:Qualitative result example 3}. The model predicts the wrong answer D - ``[0] is afraid that he will be seen.'' The correct reason is B - ``The building is on fire and he is vulnerable to it.''

\section{Conclusion}
\label{chap:conclusion}
This paper has studied the popular visual commonsense reasoning (VCR). We propose a working memory based model composed of a feature representation layer to capture multiple features containing language and objects information; a multimodal fusion layer to fuse features from language and images; an encoder layer to encode commonsense between sentences and enhance inference ability using dynamic knowledge from a dictionary unit; and a prediction layer to predict a correct response from four choices. We conduct extensive experiments on the VCR dataset to demonstrate the effectiveness of our model and present intuitive interpretation. In the future, it would be interesting to investigate multimodal feature fusion methods as well as encoding commonsense using an attention mechanism to improve the performance of VCR.


\nocite{*}
\bibliographystyle{IEEEtran}
\bibliography{ref}
\end{document}